\documentclass[10pt,twocolumn,letterpaper]{article}

\usepackage{wacv}              %

\usepackage{graphicx}
\usepackage{amsmath}
\usepackage{amssymb}
\usepackage{booktabs}

\usepackage[]{mdframed}
\usepackage{array}
\usepackage{multirow}

\usepackage[pagebackref,breaklinks,colorlinks]{hyperref}

\usepackage[capitalize]{cleveref}
\crefname{section}{Sec.}{Secs.}
\Crefname{section}{Section}{Sections}
\Crefname{table}{Table}{Tables}
\crefname{table}{Tab.}{Tabs.}

\def\wacvPaperID{630} %
\def\confName{WACV}
\def\confYear{2026}

\begin{document}

\title{Generating Accurate and Detailed Captions for High-Resolution Images}

\author{Hankyeol Lee$^{1}$, Gawon Seo$^{2}$, Kyounggyu Lee$^{1}$, Dogun Kim$^{1}$, Kyungwoo Song$^{3}$, Jiyoung Jung$^{1}$\\
$^{1}$Department of Artificial Intelligence, University of Seoul\\
$^{2}$Department of Computer Science and Engineering, POSTECH\\
$^{3}$Department of Applied Statistics, Yonsei University\\
}

\maketitle

\begin{abstract}
Vision-language models (VLMs) often struggle to generate accurate and detailed captions for high-resolution images since they are typically pre-trained on low-resolution inputs (e.g., 224×224 or 336×336 pixels).
Downscaling high-resolution images to these dimensions may result in the loss of visual details and the omission of important objects. To address this limitation, we propose a novel pipeline that integrates vision-language models, large language models (LLMs), and object detection systems to enhance caption quality.
Our proposed pipeline refines captions through a novel, multi-stage process. Given a high-resolution image, an initial caption is first generated using a VLM, and key objects in the image are then identified by an LLM. The LLM predicts additional objects likely to co-occur with the identified key objects, and these predictions are verified by object detection systems.
Newly detected objects not mentioned in the initial caption undergo focused, region-specific captioning to ensure they are incorporated.
This process enriches caption detail while reducing hallucinations by removing references to undetected objects.
We evaluate the enhanced captions using pairwise comparison and quantitative scoring from large multimodal models, along with a benchmark for hallucination detection.
Experiments on a curated dataset of high-resolution images demonstrate that our pipeline produces more detailed and reliable image captions while effectively minimizing hallucinations.
\end{abstract}

\section{Introduction}
\label{sec:intro}

Image captioning stands at the intersection of computer vision and natural language processing, aiming to generate textual narratives that capture the objects and contextual nuances in a given image.
Early methods typically employed convolutional neural networks (CNNs) and recurrent neural networks (RNNs) in an encoder-decoder setup, where the CNN extracted visual features and the RNN generated corresponding sentences~\cite{Vinyals2014ShowAT, Karpathy2014DeepVA}. While these architectures provided a strong foundation, they struggled to represent more intricate context, leading to the adoption of attention mechanisms~\cite{Xu2015ShowAA} for selectively focusing on salient image regions. Subsequently, Transformer-based models~\cite{Cornia2019MeshedMemoryTF, Herdade2019Transforming} advanced the field by enabling multi-head self-attention, allowing for both long-range dependency handling and parallel computation to produce more nuanced captions.

More recently, large vision-language models (VLMs)~\cite{dai2023instructblip, liu2023llava, openai2024gpt4technicalreport, wang2024qwen2vl} have emerged, trained on massive paired image-text datasets to learn fine-grained semantic relationships. Despite these advances, many state-of-the-art VLMs still rely on vision backbones with fixed, relatively low input resolutions~\cite{Wu2023V-star}. Downsampling large or detailed images to these resolutions risks compressing or distorting vital information~\cite{wang2024divideconquer}, such as subtle facial expressions or fine textures. This degradation can not only compromise the quality of generated captions but also weaken the model’s understanding of complex scenes, particularly where distant objects or text-based elements matter. Although several studies aim to produce more detailed image descriptions~\cite{Ge2024VisualFC, Shao2023Textual, Ramos2022SmallcapLI}, effectively handling high-resolution inputs remains an open challenge.

In this work, we introduce a training-free pipeline that leverages a vision-language model (VLM) in tandem with a large language model (LLM) and object detectors to generate richer and more accurate captions for high-resolution images. Our approach begins by using a VLM to produce an initial scene-level caption. We then harness the LLM to inspect this caption, extracting ``key objects'' and inferring additional objects likely to co-occur with the ones already identified. This inference step relies on the LLM's extensive contextual knowledge to propose potentially relevant but initially overlooked elements. Next, object detectors verify the presence of these candidate objects in the image. If previously unmentioned objects are detected, we employ a ``zoom-in'' mechanism. This involves generating detailed, object-level captions for newly verified objects by focusing on their specific regions within the image. By merging these refined descriptions with the original caption, we arrive at a more comprehensive narrative, mimicking how humans visually zoom in to inspect complex images.

\begin{figure*}[t!]
    \centering
    \includegraphics[width=17.5cm]{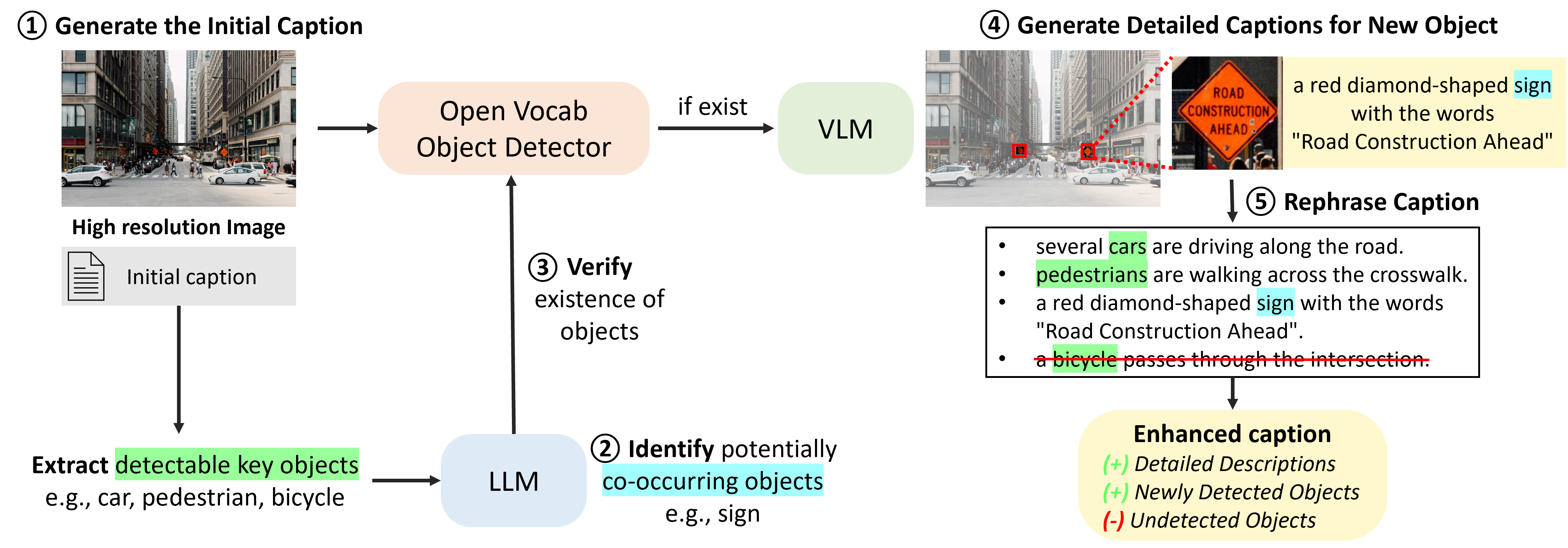}
    \caption{Overview of the caption refinement pipeline. The process begins with initial caption generation using a captioner (VLM). Next, potentially co-occurring objects are identified with the help of a large language model, followed by verifying the existence of objects using a detector. Subsequently, detailed captioning is performed to incorporate newly detected objects. Finally, the enhanced caption is generated by rephrasing with the LLM. The pipeline enhances the image caption to improve its accuracy and descriptive detail.}
    \label{fig:overview}
\end{figure*}

Conversely, if the object detectors fail to confirm certain objects mentioned in the initial caption, we remove those references from the caption. This eliminates inaccuracies and reduces hallucinations, where models include objects that are not present in the image. By refining the caption based on verified detections, we ensure the final description is both accurate and complete.

In summary, our contributions are as follows:
1) We propose a training-free pipeline that leverages an LLM and object detectors to improve caption accuracy by verifying object presence, thereby adding fine-grained details and reducing hallucinations without requiring model retraining.
2) We introduce a human-like ``zoom-in'' mechanism that generates context-rich descriptions for newly discovered objects, effectively capturing details in high-resolution images that are often overlooked by conventional VLMs.
3) As a result, our method produces comprehensively reliable and detailed image descriptions, significantly enhancing caption precision and coverage for a wide range of downstream applications.

\section{Related Works}
\subsection{Dense Captioning}
Dense captioning aims to detect multiple salient regions within an image and generate descriptive sentences for each region.
Unlike traditional image captioning, which provides a single high-level summary, dense captioning focuses on fine-grained details, capturing object interactions and attributes at a localized level.
This richer semantic information benefits tasks like visual question answering, storytelling, and assistive tools for visually impaired users.
Early works like DenseCap~\cite{Johnson-2016-DenseCap} employed region proposal mechanisms and recurrent neural networks (RNNs) to generate short captions tied to each region.
A joint inference approach~\cite{Yang2016DenseCW} further improves region classification and caption generation, enhancing consistency across proposed regions.
More recent methods have integrated advanced object detection and attention-based language models to handle high-resolution or cluttered images.
For instance, a multi-modal attention technique iteratively refines region proposals and language features, while a Transformer-based model~\cite{wang2021end} captures long-range dependencies among regions.
Furthermore, the structured attention-based method addresses the limitations of the unstructured approach by explicitly modeling spatial relationships between regions~\cite{kim2017structuredattention}. Unlike conventional attention mechanisms, which often focus on coarse and unstructured features, structured attention leverages spatial and semantic relationships between objects, enabling more accurate and context-aware caption generation.

Recent approaches have explored integrating dense captioning with open-vocabulary detection~\cite{zareian2021openvocabularyobjectdetection} to address the challenges of handling diverse object categories and generating detailed descriptions. For example, CapDet~\cite{long2023capdet} unifies dense captioning and open-vocabulary detection pretraining process into a single framework, enabling the model to detect and recognize both common objects and novel categories while generating region-grounded captions. These developments highlight the potential ability of dense captioning systems to adapt to real-world scenarios with more flexible and comprehensive capabilities.

\subsection{Addressing Hallucination in Image Captioning}
Hallucinations in image captions mainly stem from the misalignment between image features and linguistic priors learned from the dataset.
Research has identified two primary types of hallucination: object hallucination, where nonexistent objects are mentioned, and attribute hallucination, where incorrect properties are assigned to objects.
This phenomenon poses significant challenges for deploying captioning systems where reliability is crucial in real-world applications.
Building upon early CNN-RNN architectures~\cite{Vinyals2014ShowAT}, the integration of attention mechanisms~\cite{Xu2015ShowAA} and object detection modules~\cite{Anderson2017BottomUpAT} has improved the model's ability to ground generated captions in visual content.
Recent approaches have made significant strides in addressing hallucinations through multiple strategies.
Constrained decoding methods~\cite{Lu-2018-Neural} and uncertainty-aware generation techniques~\cite{Zhengcong-2023-Uncertainty} provide explicit control over the caption generation process, reducing the likelihood of hallucinated content.
The emergence of large-scale vision-language pre-training has revolutionized the field, with models like CLIP~\cite{radford2021clip} and VisualBERT~\cite{li2019visualbert} significantly improved the alignment between visual and linguistic representations.

Despite these advancements, many captioning pipelines still rely on initially generated text, which can lack fine-grained details or introduce factual errors (i.e., hallucinations), thereby limiting both the effectiveness and reliability of large-scale pre-training.
Accordingly, recent work has explored ways to refine these initial captions, either by merging existing synthetic data through large language models~\cite{yu2024capsfusion} or by verifying them using post hoc fact-checking~\cite{ge2024visualfactchecker}, thereby reducing hallucinations and incomplete descriptions to produce more reliable and contextually enriched captions.
Additionally, recent work has introduced novel evaluation metrics specifically designed to detect and measure hallucination~\cite{li2023pope}, enabling more targeted improvements in model development.
These advances collectively contribute to more reliable and factual image captioning systems, though challenges remain in achieving human-level accuracy and consistency.

Furthermore, the quality of text embeddings—the numerical representations of textual inputs—plays a crucial role in aligning language with visual content in image captioning models. Well-structured embeddings help models accurately associate words with corresponding visual elements, leading to more semantically coherent and faithful captions. On the other hand, weak or misaligned embeddings may cause hallucinated or vague descriptions. Recent work, such as CLIP~\cite{radford2021clip} and CapsFusion~\cite{yu2024capsfusion}, underscores the significance of robust image-text representation learning, showing that improvements in embedding quality can enhance caption reliability, particularly in high-resolution or complex scenes.

\subsection{High-Resolution Image Processing in VLMs}
A significant challenge for many vision-language models (VLMs) is the effective processing of high-resolution images. Most VLMs are pre-trained with relatively low-resolution image inputs (e.g., 224x224 or 336x336 pixels), creating a "resolution curse" where vital details in high-resolution images are lost during the downsampling process. To address this, several strategies have been proposed.

One dominant approach is patch-based processing. These methods divide a high-resolution image into multiple smaller patches and feed them to the VLM. For instance, \cite{carvalho-martins-2025-efficient} introduces an efficient architecture that uses two sets of LoRA adapters to process features from a low-resolution global image and high-resolution local patches simultaneously, enhancing the model's ability to perceive both context and detail. While effective, such methods often require architectural modifications and specialized multi-stage training procedures.

Another emerging direction is inspired by the human visual system, employing Fovea-based or multi-resolution approaches. These methods mimic the human eye's ability to perceive a scene with a high-resolution central region and a lower-resolution peripheral view. For example, \cite{gizdov2025seeing} apply foveated sampling to large multimodal models (LMMs), demonstrating that this biologically inspired technique can create more human-like visual representations and improve performance on tasks like visual question answering without needing to process the entire image at maximum resolution.

\section{Method}
We introduce a training-free pipeline designed to refine image captions, improving both their accuracy and descriptive detail. The process begins with generating an initial caption based on visual input, which is then progressively enriched through object detection, co-occurrence analysis, and rephrasing. By leveraging both the vision-language model (VLM) and the large language model (LLM), we ensure that the final caption reflects a comprehensive understanding of the image, incorporating key objects, their relationships, and precise spatial information. The overview of our method is illustrated in Fig.~\ref{fig:overview}.

\subsection{Generating the Initial Caption} \label{sec:method-1}
Since our pipeline focuses on refining captions, the process begins with generating an initial caption using VLMs~\cite{dai2023instructblip, liu2023improvedllava, wang2024qwen2vl}.

When provided a high-resolution image as input, the VLM often overlooks fine details or generates hallucinated content. To generate the initial caption, we input the image into the model with a simple prompt, ``\texttt{Describe this image in detail.}''

Next, we employ an LLM, specifically GPT-4o~\cite{hurst2024gpt4o}, to extract key objects from the initial caption, that are suitable for the subsequent object detection stage. To detect key objects and identify potentially co-occurring objects, we use the prompt as shown in Fig.~\ref{fig:overview}.
The \textbf{input} for this stage is a high-resolution image. The \textbf{outputs} are twofold: the initial descriptive caption (as a string) and a structured list of key object names identified within it.

\subsection{Identifying Potentially Co-occurring Objects}

A key objective of this step is to identify objects present in the image but omitted from the initial caption. These missing objects are crucial for creating a more comprehensive and accurate image description. To achieve this, we leverage the extensive knowledge embedded in the LLM.
This step transcends simple visual analysis by leveraging the LLM's vast world knowledge to reason about plausible scene compositions. The core task is to use the initial key objects as contextual anchors to infer other items that are likely present but were omitted from the initial caption. This approach is analogous to human common-sense reasoning, enriching the scene's context beyond the explicitly mentioned elements.

To implement this, the process starts by inputting the list of key objects into the LLM. We use a specially crafted prompt designed to encourage the model to propose objects that frequently co-occur with the given anchors. For example, if the key objects are ``table'' and ``chair'', the LLM is prompted to suggest related items commonly found in such a setting, like a ``lamp'', ``books'', or a ``cup''. This stage takes the list of key objects as input and outputs an expanded candidate list of objects, which includes both the original and the newly proposed items.

This stage takes the list of key objects as \textbf{input} and \textbf{outputs} an expanded candidate list of objects, which includes both the original and the newly proposed items.

\subsection{Verifying the Existence of Objects}

To generate accurate captions, it is essential to verify whether the objects described actually exist in the image. For this purpose, we employ three object detectors~\cite{Liu2023GroundingDINO, Cheng2024YOLOWorld, Minderer2023OWLv2}. Since object detectors are generally trained on higher-resolution images compared to VLMs, they offer improved object detection accuracy.
We use an ensemble of three detectors to ensure robustness, as a single detector may have inherent biases from its training data. By combining models with proven open-vocabulary capabilities, we mitigate individual weaknesses and achieve more reliable object verification across diverse scenarios.

This stage filters the candidate object list—which contains both items from the initial caption and new proposals from the LLM—against the visual evidence in the image. Based on the detection results, this list is refined into a final set of visually confirmed objects. Specifically, candidate objects successfully located by the detectors are retained, while those that cannot be verified are discarded. This refined list serves as the foundation for the subsequent captioning steps.

An object is considered newly detected if the combined confidence scores from all three detectors reach 0.5 or higher. During this process, objects are regarded as identical if their Intersection over Union (IoU) score is 0.7 or higher across detection results. Conversely, an object is removed from the initial caption if it is not detected by any of the three detectors.
The \textbf{inputs} to this verification stage are the high-resolution image and the candidate object list. The \textbf{output} is a refined list of visually confirmed objects, each associated with its bounding box coordinates, which serves as the visual ground truth for the final rephrasing step.

\subsection{Generating Detailed Captions for New Objects}

To enhance descriptive quality, we adopt a detailed captioning approach, addressing instances where newly detected objects were not adequately represented in the initial description. By focusing on these objects, we ensure a more elaborate and comprehensive image description.

To achieve this, we crop the bounding boxes of newly detected objects and re-input them into the VLM. This process mimics the way a person zooms in on specific areas of a high-resolution image to observe finer details. Importantly, this approach is applied only to objects absent from the initial caption to minimize computational cost, based on the assumption that previously described objects are already sufficiently detailed. For caption generation, we use the same model and prompt as outlined in Section~\ref{sec:method-1}.
The \textbf{input} for this stage is a set of cropped image regions for newly verified objects. The \textbf{output} is a collection of corresponding detailed textual descriptions for each region.

\subsection{Rephrasing the Final Caption}
In the final step, we rephrase the initial caption using detailed object detection results and captions gathered from prior processing steps. This is achieved with GPT-4o, which excels in generating coherent and context-aware text. While the image itself is not processed directly, the model utilizes structured prompts containing descriptions, spatial relationships, and detection results. %

The purpose of caption rephrasing in our pipeline is twofold: to remove inaccurate phrases related to undetected objects from the original caption, and to incorporate newly detected objects with more detailed and context-aware descriptions. Rather than simply appending or deleting individual sentences, we rephrase the entire caption to ensure that the resulting text remains natural, coherent, and easy to understand. This holistic editing approach preserves the fluency of the narrative while enhancing both the accuracy and richness of the final caption.

Newly detected objects are incorporated into the caption seamlessly, along with their spatial contexts, described in relative terms such as ``on the left,'' ``in the foreground,'' or ``near the center.'' Their detection details are formatted as: \texttt{\{label\}: \{caption\} at coordinates (x\_min: \{box[0]\}, y\_min: \{box[1]\}, x\_max: \{box[2]\}, y\_max: \{box[3] \newline \})}, enabling precise integration. For instance, a newly detected ``lamp'' might be expressed as ``A lamp is positioned to the right of the table.''
We also ensure that objects mentioned in the initial caption but not detected by any of the object detectors are removed while preserving other descriptive attributes such as color, shape, and context. This approach enhances the caption’s accuracy without introducing unnecessary changes, allowing it to maintain its natural flow.
The full templates for all prompts used in our pipeline are available in the supplementary material.

The resulting enhanced caption is comprehensive and contextually enriched, reflecting the image more accurately. Each object is described in a way that integrates naturally into the narrative, while spatial contexts add depth. Non-existent or irrelevant objects are carefully excluded, ensuring the output is refined and precise.
The \textbf{inputs} for this final synthesis stage are: (1) the initial caption, (2) the verified list of objects with their spatial coordinates, and (3) the new detailed descriptions. The final \textbf{output} is a single, coherent, and contextually rich enhanced caption.

\section{Experiments}
In Section~\ref{sec:experiments-1}, we provide detailed experimental information, including an overview of the models utilized, the dataset used, and the filtering criteria applied to construct its subsets. These details provide the foundation for reproducibility and a comprehensive understanding of our experimental setup.

We then present two experiments and their results to demonstrate the effectiveness of our method. Specifically, we either enrich the initial captions with new information or remove existing erroneous information. These enhancements aim to improve the informativeness and reliability of the captions while preserving their conciseness.
We evaluate the improvement in overall caption quality using a large multimodal model in Section~\ref{sec:experiments-2}. Additionally, to verify the removal of inaccurate information, we conduct an evaluation using the hallucination benchmark, POPE~\cite{li2023pope}, as detailed in Section~\ref{sec:experiments-3}. By combining these evaluations, we aim to provide a comprehensive analysis of the method’s impact across multiple perspectives.

\subsection{Experimental Details}
\label{sec:experiments-1}

\subsubsection{Models}
We use three vision-language models (VLMs) to generate the initial caption: InstructBLIP~\cite{dai2023instructblip}, LLaVA-v1.5~\cite{liu2023improvedllava}, and Qwen2-VL~\cite{wang2024qwen2vl}. InstructBLIP resizes images to 224×224, LLaVA-v1.5 resizes images to 336×336, and Qwen2-VL employs a dynamic resolution mechanism, making it robust across various resolutions. We select these models to examine how their performance varies according to their input image resizing dimensions.

To extract key objects, rephrase captions, and perform other tasks within the pipeline that involve large language models, we primarily rely on GPT-4o~\cite{hurst2024gpt4o}, unless explicitly specified otherwise.

For refining captions, we integrate three object detectors: GroundingDINO~\cite{Liu2023GroundingDINO}, YOLO-World~\cite{Cheng2024YOLOWorld}, and OWLv2~\cite{Minderer2023OWLv2}. These detectors are all trained with an open-vocabulary approach, allowing them to adapt to a diverse range of scenarios.

In evaluating caption quality, we use the state-of-the-art large multimodal model, LLaMA-3.2-Vision-Instruct~\cite{grattafiori2024llama3herdmodels}. This open-source model demonstrates state-of-the-art performance across various vision and language tasks while being free from cost constraints. It can process images up to 1120×1120 pixels, resizing larger images to fit within these dimensions.

Finally, we use GPT-4o to evaluate hallucination in captions. The POPE~\cite{li2023pope} benchmark we employed was originally designed as a polling-based approach to evaluate hallucination in vision-language models. Since evaluating hallucination in captions requires an additional model, we utilize GPT-4o to generate answers.

\subsubsection{Dataset}
We discuss the performance degradation observed in vision-language models when high-resolution images are used as input. Consequently, we emphasize the need for performance evaluation on high-resolution images. To address this, we filter 4K resolution (3840×2160 pixels) images from the Objects365 dataset~\cite{Shao2019Objects365}, focusing specifically on images with complex contexts.

Specifically, we select images that meet the following criteria: containing at least 15 unique object classes, 10 small objects (each occupying less than 1\% of the image size), and a minimum of 5 people. A total of 266 images satisfy these conditions, and we use them to evaluate the performance of our method.

\begin{figure}[t!]
  \centering
  \includegraphics[width=8cm]{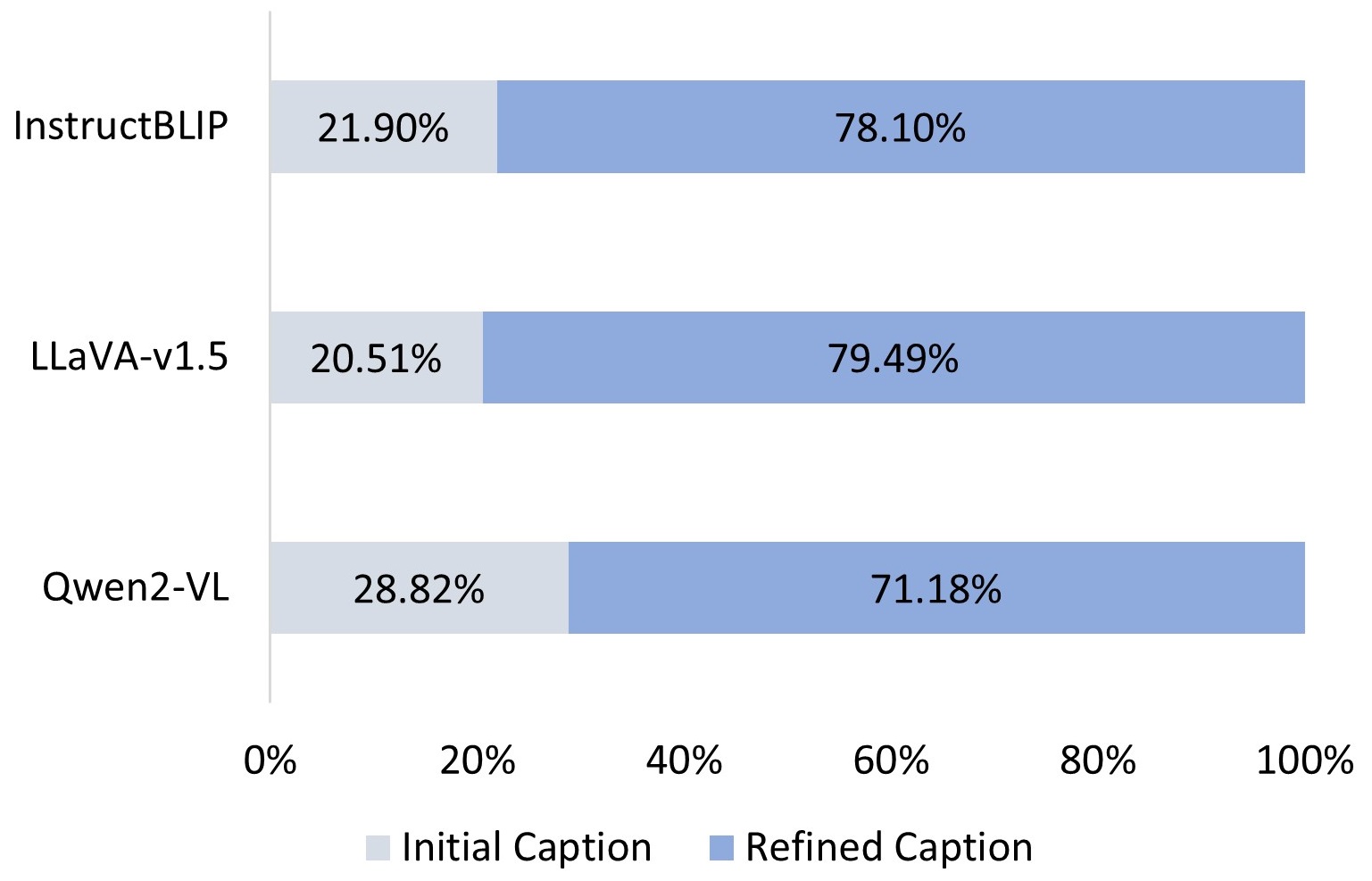}
  \caption{Winning rates for pairwise caption comparison.}
  \label{fig:pairwise}
\end{figure}

\begin{table}[t!]
\centering
\begin{tabular}{@{}lcc@{}}
\toprule
\multicolumn{1}{c}{\textbf{Model}} & \textbf{Score} & \textbf{Improvement} \\ \midrule
InstructBLIP & 0.6344 & -- \\
+ Ours       & 0.6952 & {\color[rgb]{0.2,0.6,0.2}+9.59\%} \\ \midrule
LLaVA-v1.5   & 0.6785 & -- \\
+ Ours       & 0.7304 & {\color[rgb]{0.2,0.6,0.2}+7.66\%} \\ \midrule
Qwen2-VL     & 0.8260 & -- \\
+ Ours       & 0.8398 & {\color[rgb]{0.2,0.6,0.2}+1.68\%} \\ \bottomrule
\end{tabular}
\caption{Quantitative scores of the captions. Our method consistently enhances caption quality across all evaluated captioners.}
\label{tab:quantitative-score}
\end{table}

\begin{figure*}[t!]
  \centering
  \includegraphics[width=17cm]{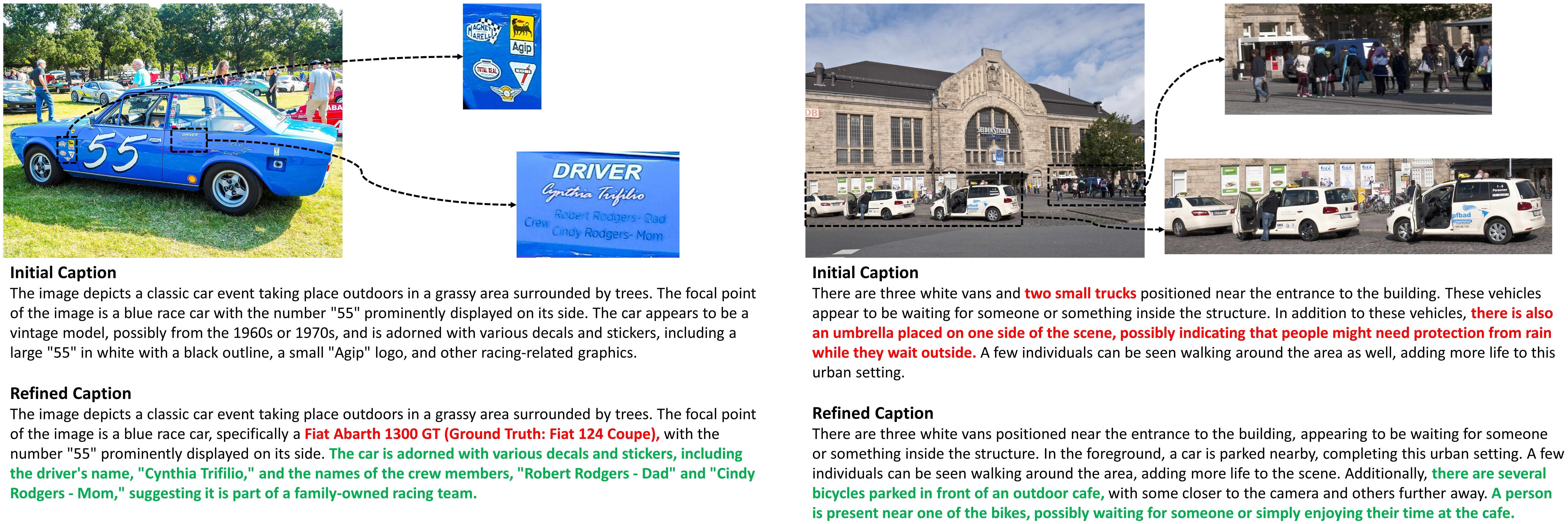}
  \caption{Qualitative comparison of initial and enhanced caption. The green text represents newly added information, and the red text indicates hallucination or incorrect information. Our method generates detailed and reliable captions.}
  \label{fig:qualitative}
\end{figure*}

\begin{table*}[t]
\centering
\begin{tabular}{@{}c l cccc@{}}
\toprule
\textbf{POPE} &
  \multicolumn{1}{c}{\textbf{Model}} &
  \multicolumn{1}{c}{\textbf{Accuracy}} &
  \multicolumn{1}{c}{\textbf{Precision}} &
  \multicolumn{1}{c}{\textbf{Recall}} &
  \multicolumn{1}{c}{\textbf{F1 Score}} \\ \midrule
\multirow{6}{*}[-1.75em]{\centering \textit{Random}} &
  InstructBLIP &
  0.4893 &
  0.4467 &
  0.0890 &
  0.1484 \\
 &
  + Ours &
  \parbox[c]{2.5cm}{\centering 0.5132 \\ \textcolor[rgb]{0.2, 0.6, 0.2}{(+4.87\%)}} &
  \parbox[c]{2.5cm}{\centering 0.5546 \\ \textcolor[rgb]{0.2, 0.6, 0.2}{(+24.16\%)}} &
  \parbox[c]{2.5cm}{\centering 0.1336 \\ \textcolor[rgb]{0.2, 0.6, 0.2}{(+50.15\%)}} &
  \parbox[c]{2.5cm}{\centering 0.2153 \\ \textcolor[rgb]{0.2, 0.6, 0.2}{(+45.10\%)}} \\ \cmidrule(l){2-6} 
 &
  LLaVA-v1.5 &
  0.4895 &
  0.4582 &
  0.1155 &
  0.1845 \\
 &
  + Ours &
  \parbox[c]{2.5cm}{\centering \textbf{0.5154} \\ \textcolor[rgb]{0.2, 0.6, 0.2}{(+5.30\%)}} &
  \parbox[c]{2.5cm}{\centering \textbf{0.5601} \\ \textcolor[rgb]{0.2, 0.6, 0.2}{(+22.23\%)}} &
  \parbox[c]{2.5cm}{\centering \textbf{0.1436} \\ \textcolor[rgb]{0.2, 0.6, 0.2}{(+24.31\%)}} &
  \parbox[c]{2.5cm}{\centering \textbf{0.2286} \\ \textcolor[rgb]{0.2, 0.6, 0.2}{(+23.88\%)}} \\ \cmidrule(l){2-6} 
 &
  Qwen2-VL &
  0.4943 &
  0.4704 &
  0.0895 &
  0.1504 \\
 &
  + Ours &
  \parbox[c]{2.5cm}{\centering 0.5073 \\ \textcolor[rgb]{0.2, 0.6, 0.2}{(+2.62\%)}} &
  \parbox[c]{2.5cm}{\centering 0.5349 \\ \textcolor[rgb]{0.2, 0.6, 0.2}{(+13.70\%)}} &
  \parbox[c]{2.5cm}{\centering 0.1118 \\ \textcolor[rgb]{0.2, 0.6, 0.2}{(+24.90\%)}} &
  \parbox[c]{2.5cm}{\centering 0.1849 \\ \textcolor[rgb]{0.2, 0.6, 0.2}{(+22.97\%)}} \\ \midrule
\multirow{6}{*}[-1.75em]{\centering \textit{Popular}} &
  InstructBLIP &
  0.4902 &
  0.4582 &
  0.1068 &
  0.1732 \\
 &
  + Ours &
  \parbox[c]{2.5cm}{\centering 0.5098 \\ \textcolor[rgb]{0.2, 0.6, 0.2}{(+3.99\%)}} &
  \parbox[c]{2.5cm}{\centering 0.5373 \\ \textcolor[rgb]{0.2, 0.6, 0.2}{(+17.27\%)}} &
  \parbox[c]{2.5cm}{\centering 0.1406 \\ \textcolor[rgb]{0.2, 0.6, 0.2}{(+31.68\%)}} &
  \parbox[c]{2.5cm}{\centering 0.2229 \\ \textcolor[rgb]{0.2, 0.6, 0.2}{(+28.71\%)}} \\ \cmidrule(l){2-6} 
 &
  LLaVA-v1.5 &
  0.4885 &
  0.4588 &
  0.1286 &
  0.2008 \\
 &
  + Ours &
  \parbox[c]{2.5cm}{\centering \textbf{0.5108} \\ \textcolor[rgb]{0.2, 0.6, 0.2}{(+4.57\%)}} &
  \parbox[c]{2.5cm}{\centering 0.5372 \\ \textcolor[rgb]{0.2, 0.6, 0.2}{(+17.08\%)}} &
  \parbox[c]{2.5cm}{\centering \textbf{0.1559} \\ \textcolor[rgb]{0.2, 0.6, 0.2}{(+21.25\%)}} &
  \parbox[c]{2.5cm}{\centering \textbf{0.2417} \\ \textcolor[rgb]{0.2, 0.6, 0.2}{(+20.32\%)}} \\ \cmidrule(l){2-6} 
 &
  Qwen2-VL &
  0.4944 &
  0.4734 &
  0.1008 &
  0.1661 \\
 &
  + Ours &
  \parbox[c]{2.5cm}{\centering 0.5104 \\ \textcolor[rgb]{0.2, 0.6, 0.2}{(+3.24\%)}} &
  \parbox[c]{2.5cm}{\centering \textbf{0.5447} \\ \textcolor[rgb]{0.2, 0.6, 0.2}{(+15.05\%)}} &
  \parbox[c]{2.5cm}{\centering 0.1268 \\ \textcolor[rgb]{0.2, 0.6, 0.2}{(+25.86\%)}} &
  \parbox[c]{2.5cm}{\centering 0.2057 \\ \textcolor[rgb]{0.2, 0.6, 0.2}{(+23.84\%)}} \\ \midrule
\multirow{6}{*}[-1.75em]{\centering \textit{Adversarial}} &
  InstructBLIP &
  0.4886 &
  0.4529 &
  0.1098 &
  0.1767 \\
 &
  + Ours &
  \parbox[c]{2.5cm}{\centering 0.5095 \\ \textcolor[rgb]{0.2, 0.6, 0.2}{(+4.28\%)}} &
  \parbox[c]{2.5cm}{\centering 0.5352 \\ \textcolor[rgb]{0.2, 0.6, 0.2}{(+18.17\%)}} &
  \parbox[c]{2.5cm}{\centering 0.1446 \\ \textcolor[rgb]{0.2, 0.6, 0.2}{(+31.74\%)}} &
  \parbox[c]{2.5cm}{\centering 0.2277 \\ \textcolor[rgb]{0.2, 0.6, 0.2}{(+28.85\%)}} \\ \cmidrule(l){2-6} 
 &
  LLaVA-v1.5 &
  0.4915 &
  0.4698 &
  0.1323 &
  0.2065 \\
 &
  + Ours &
  \parbox[c]{2.5cm}{\centering \textbf{0.5147} \\ \textcolor[rgb]{0.2, 0.6, 0.2}{(+4.71\%)}} &
  \parbox[c]{2.5cm}{\centering 0.5501 \\ \textcolor[rgb]{0.2, 0.6, 0.2}{(+17.11\%)}} &
  \parbox[c]{2.5cm}{\centering \textbf{0.1609} \\ \textcolor[rgb]{0.2, 0.6, 0.2}{(+21.58\%)}} &
  \parbox[c]{2.5cm}{\centering \textbf{0.2490} \\ \textcolor[rgb]{0.2, 0.6, 0.2}{(+20.59\%)}} \\ \cmidrule(l){2-6} 
 &
  Qwen2-VL &
  0.4895 &
  0.4502 &
  0.0953 &
  0.1572 \\
 &
  + Ours &
  \parbox[c]{2.5cm}{\centering 0.5115 \\ \textcolor[rgb]{0.2, 0.6, 0.2}{(+4.51\%)}} &
  \parbox[c]{2.5cm}{\centering \textbf{0.5507} \\ \textcolor[rgb]{0.2, 0.6, 0.2}{(+22.32\%)}} &
  \parbox[c]{2.5cm}{\centering 0.1248 \\ \textcolor[rgb]{0.2, 0.6, 0.2}{(+31.01\%)}} &
  \parbox[c]{2.5cm}{\centering 0.2035 \\ \textcolor[rgb]{0.2, 0.6, 0.2}{(+29.44\%)}} \\ \bottomrule
\end{tabular}%
\caption{POPE benchmark results for initial and enhanced captions, comparing baseline models with and without our enhancement across Random, Popular, and Adversarial settings. Metrics include Accuracy, Precision, Recall, and F1 Score. ``+ Ours'' rows highlight performance improvements, with percentage gains in green, demonstrating the robustness and effectiveness of our method in diverse settings.}
\label{tab:POPE}
\end{table*}

\subsection{Evaluation of Caption Quality}
\label{sec:experiments-2}
Recently, various metrics for evaluating captions have been proposed. Broadly, these metrics can be categorized into reference-based methods~\cite{Jiang-2019-TIGEr, Wada2024Polos}, which require reference captions, and reference-free methods~\cite{Lee-2021-UMIC, Hu-2023-InfoMetIC, Sarto2023PositiveAugmentedCL, Hessel-2021-CLIPScore} which evaluate captions without reference captions. Since the dataset used in our experiments does not include reference captions, we adopt a reference-free evaluation approach.
However, a limitation of existing reference-free is that most studies~\cite{Hu-2023-InfoMetIC, Sarto2023PositiveAugmentedCL, Hessel-2021-CLIPScore} rely on CLIP~\cite{radford2021clip} as the backbone model. While CLIP has demonstrated remarkable performance, its image encoder has a relatively low resolution (224×224), making it unsuitable for evaluating the quality of captions generated from the high-resolution images used in our experiments.

To address this limitation, we employ an open-source large multimodal model (LMM), LLaMA-3.2-Vision-Instruct~\cite{grattafiori2024llama3herdmodels}, which can handle higher-resolution images (up to 1120×1120). Following recent works~\cite{Chan-2023-CLAIR, Lee-2024-FLEUR, Ge2024VisualFC} that demonstrate the effectiveness of LMMs in evaluation tasks, our approach leverages these models to provide more comprehensive evaluation capabilities while maintaining reliable assessment of caption quality at a fraction of the cost of human evaluation.

We design two experiments to evaluate our approach. In the first experiment, we provide the model with an image and its corresponding captions---an initial caption and an enhanced caption---allowing it to choose which is better. This evaluation is conducted pairwise, where each caption pair is directly compared to determine which is superior. Notably, we focus on two primary aspects: correctness and detail. The LMM evaluates whether the enhanced caption demonstrates improvement over the initial one and justifies its evaluation. The experimental results are presented in Fig.~\ref{fig:pairwise}. The prompts used in the experiments are based on~\cite{Ge2024VisualFC}.
To ensure reliability, we repeat the same experiment five times and report the average scores. Our method shows a high winning rate across all three captioners used in the experiment.

In the second experiment, we input an image and its caption into the model, which outputs a quantitative score ranging from 0.0 to 1.0. The prompt used in this experiment is adapted from \cite{Lee-2024-FLEUR}.
The experimental results, obtained by repeating the experiment five times and reporting the average scores, are presented in Table~\ref{tab:quantitative-score}. As in the previous experiment, our method shows improved performance across all three captioners.
Interestingly, Qwen2-VL, which has the highest initial score, shows the smallest percentage improvement. This is likely because its dynamic resolution mechanism makes it inherently more robust, leaving less room for enhancement. This observation underscores our method's significant value, particularly for VLMs more prone to information loss from low-resolution processing.

Furthermore, the results of the qualitative evaluation are shown in Fig.~\ref{fig:qualitative}. This figure provides an intuitive understanding of the effectiveness of our method.

\subsection{Evaluation of Hallucination in Caption}
\label{sec:experiments-3}
We not only add new information to the captions but also effectively eliminate hallucinations, thereby removing incorrect information. To evaluate the effectiveness in this aspect, we utilize the hallucination benchmark POPE~\cite{li2023pope}. Originally proposed as a polling-based benchmark for assessing hallucinations in vision-language models, we assess hallucinations in captions by inputting the generated captions and POPE questions into GPT-4o~\cite{hurst2024gpt4o}, a large language model to generate answers.

Specifically, POPE employs three sampling strategies: random sampling, which involves randomly selecting objects that do not appear in the image; popular sampling, which selects the top-k most frequent objects across the entire image dataset that are absent in the current image; and adversarial sampling, which ranks all objects based on their co-occurrence frequencies with the ground-truth objects and selects the top-\textit{k} frequent ones that are not present in the image~\cite{li2023pope}.

We present the experimental results in Table~\ref{tab:POPE}. This experiment is also conducted five times, and the average scores are reported. In the POPE evaluation, we measure accuracy, precision, recall, and F1 score, all of which are significantly improved by our approach. Notably, unlike previous experiments, the performance improvements with Qwen2-VL~\cite{wang2024qwen2vl} are consistently significant, further demonstrating that our method contributes to the generation of more reliable captions.

\section{Conclusion}
We proposed a method to generate more detailed and reliable image captions by identifying and incorporating objects present in images that were omitted in the original captions. We recognized that VLMs trained on low-resolution images may miss important details in high-resolution images.

To address this, we combined LLMs and object detection models to detect unmentioned objects. By generating detailed captions for these newly detected objects and appending them to the original caption, we enhanced the richness and accuracy of the textual descriptions while naturally mitigating the hallucination problem. We evaluated the effectiveness of our enhanced captions using a large multimodal model, confirming qualitative and quantitative improvements.

Our experimental results demonstrated that the proposed methods effectively enhance the performance and robustness of VLMs in captioning tasks, particularly for high-resolution images, by refining caption details and integrating omitted information. By bridging the semantic gaps between visual and textual modalities, the enhanced captions provide more accurate, informative, and contextually rich outputs.

Despite the strengths of our approach, certain limitations remain. The pipeline's performance is dependent on the capabilities of the chosen object detectors, and its sequential nature can introduce latency. Future work could explore integrating a more efficient, single open-vocabulary model to replace the detector ensemble, potentially reducing computational overhead. In addition, employing more powerful pretrained VLMs and LLMs could further enhance the quality of the generated captions within our framework. As image captioning serves as a core mechanism across numerous vision-language applications, our framework has the potential to refine task-specific accuracy in scenarios such as multimodal retrieval tasks, including both text-to-image and image-to-text retrieval, text-guided image generation, and visual content description based on textual queries. Notably, it may enable these tasks to be effectively handled using high-resolution images, rather than being constrained to low-resolution inputs as in prior works. Furthermore, extending this pipeline to video captioning, where object co-occurrence and temporal consistency are crucial, presents a promising research direction. To achieve this, the pipeline must be adapted to leverage pretrained models specialized for video or other temporal modalities. Such adaptation could pave the way for broader multimodal integration and support more coherent and accurate caption generation in temporally dynamic settings.

{\small
\bibliographystyle{ieeenat_fullname}
\bibliography{main}
}

\end{document}